\documentclass{article}

\usepackage[preprint]{neurips_2025}

\usepackage[utf8]{inputenc}   
\usepackage[T1]{fontenc}      

\usepackage{microtype}        
\usepackage{xcolor}           
\usepackage{listings}         

\lstdefinelanguage{json}{
  basicstyle=\ttfamily,
  numbers=left,
  numberstyle=\tiny\color{gray},
  stepnumber=1,
  numbersep=8pt,
  showstringspaces=false,
  breaklines=true,
  frame=lines,
  literate=
   *{0}{{{\color{blue}0}}}{1}
    {1}{{{\color{blue}1}}}{1}
    {2}{{{\color{blue}2}}}{1}
    {3}{{{\color{blue}3}}}{1}
    {4}{{{\color{blue}4}}}{1}
    {5}{{{\color{blue}5}}}{1}
    {6}{{{\color{blue}6}}}{1}
    {7}{{{\color{blue}7}}}{1}
    {8}{{{\color{blue}8}}}{1}
    {9}{{{\color{blue}9}}}{1}
    {:}{{{\color{red}:}}}{1}
    {,}{{{\color{red},}}}{1}
    {"}{{{\color{orange}"}}}{1}
}

\usepackage{amsmath}          
\usepackage{amsfonts}         
\usepackage{amssymb}          
\usepackage{nicefrac}         

\usepackage{algorithm}        
\usepackage{algpseudocode}    

\usepackage{booktabs}         
\usepackage{graphicx}         

\usepackage{hyperref}         
\usepackage{url}              

\usepackage{geometry}         

\definecolor{commentcolor}{RGB}{0,100,0}
\definecolor{keywordcolor}{RGB}{0,0,255}
\definecolor{stringcolor}{RGB}{255,0,0}

\algrenewcommand\algorithmicrequire{\textbf{Input:}}
\algrenewcommand\algorithmicensure{\textbf{Output:}}

\usepackage{amsthm}

\title{Federation of Agents: A Semantics-Aware Communication Fabric for Large-Scale Agentic AI}

\author{%
Lorenzo~Giusti\thanks{CERN, Corresponding to: lorenzo.giusti@cern.ch} \And
Ole~Anton~Werner \And
Riccardo~Taiello \And
Matilde~Carvalho~Costa \And
Emre~Tosun \And
Andrea~Protani \And
Marc~Molina \And
Rodrigo~Lopes~de~Almeida \And
Paolo~Cacace \And
Diogo~Reis~Santos \And
Luigi~Serio\\[1ex]\and
\makebox[\textwidth][c]{CERN, Geneva, Switzerland}
}

\begin{document}

\maketitle

\vspace{-15pt}
\begin{abstract}
We present \emph{Federation of Agents} (FoA), a distributed orchestration framework that transforms static multi-agent coordination into dynamic, capability-driven collaboration. FoA introduces \emph{Versioned Capability Vectors} (VCVs): machine-readable profiles that make agent capabilities searchable through semantic embeddings, enabling agents to advertise their capabilities, cost, and limitations. Our architecture combines three key innovations: (1) \emph{semantic routing} that matches tasks to agents over sharded HNSW indices while enforcing operational constraints through cost-biased optimization, (2) \emph{dynamic task decomposition} where compatible agents collaboratively break down complex tasks into DAGs of subtasks through consensus-based merging, and (3) \emph{smart clustering} that groups agents working on similar subtasks into collaborative channels for $k$-round refinement before synthesis. Built on top of MQTT's publish-subscribe semantics for scalable message passing, FoA achieves sub-linear complexity through hierarchical capability matching and efficient index maintenance. Evaluation on HealthBench shows 13$\times$ improvement over single-model baselines, with clustering-enhanced collaboration particularly effective for complex reasoning tasks requiring multiple perspectives. The system scales horizontally while maintaining consistent performance, demonstrating that semantic orchestration with structured collaboration can unlock the collective intelligence of heterogeneous federations of AI agents.
\end{abstract}

\begin{figure}[!htb]
    \centering
    \vspace{-7pt}
    \includegraphics[width=0.7\linewidth]{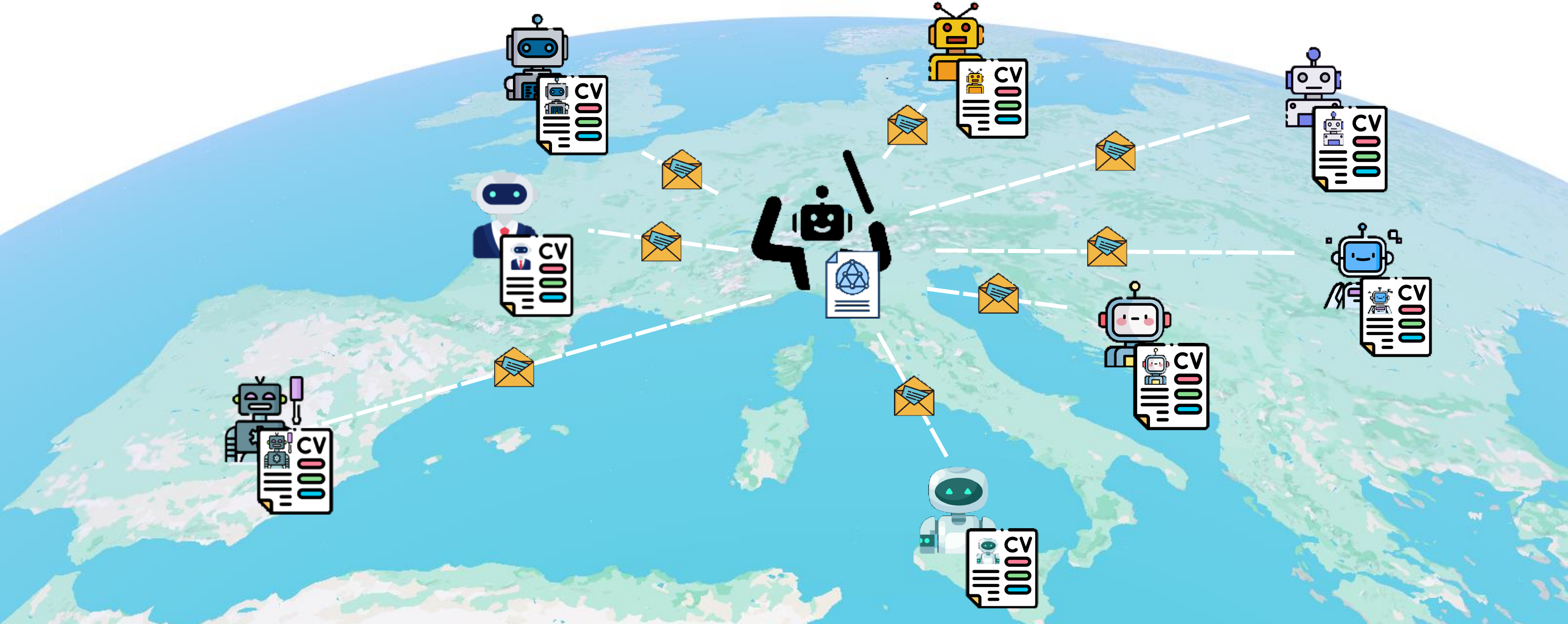}
    \caption{\textbf{Federation of Agents.} Heterogeneous agents advertise \emph{Versioned Capability Vectors} containing semantic embeddingsof agents' profiles. The orchestrator ingests these capability updates, and when a complex task arrives, it performs semantic routing to the most compatible agents or clusters. Envelopes and dashed arrows depict MQTT publish/subscribe exchanges for assignments, capability updates, and intermediate artifacts. Agents collaborate and return \texttt{TASK\_COMPLETE} signals with subtask outputs; the orchestrator merges them along the task DAG to synthesize the solution.}
    \label{fig:foe}
    \vspace{-10pt}
\end{figure}

\newpage
\section{Introduction}\label{sec:intro}

The landscape of artificial intelligence has evolved from single AI models to networks of specialized agents that plan, coordinate, and act over extended horizons~\cite{sapkota2025ai,schneider2025generative,gridach2025agentic}. This shift toward \emph{agentic AI systems} represents a fundamental change in how we approach complex problem-solving with AI: rather than relying on a single model to handle all aspects of a task, we now orchestrate collections of specialized agents that can decompose tasks, maintain persistent context, and coordinate their efforts through structured communication towards a common goal. However, current agentic AI systems mainly rely on manually curated integrations and topic-based routing~\cite{shen2023hugginggpt,yue2025masrouter}, posing constraints on scalability as the heterogeneity of agents grows, and coordination complexity increases, limiting scalability and not addressing the fundamental operational question: \emph{who can do what, at what cost, and under which policy constraints?}; preventing the realization of the "Internet of Agents" vision~\cite{wang2025internet}. To address this, we introduce \emph{Federation of Agents} (FoA), a semantics-aware communication fabric that transforms agent coordination from static, topic-based routing to dynamic, capability-driven orchestration. At its core, FoA enables agents, tools, and data stores to advertise \emph{Versioned Capability Vectors} (VCVs): machine-readable profiles that capture functional capabilities, performance characteristics, operational constraints, and security labels in a structured format.

\paragraph{Problem Statement and Challenges.} Despite rapid progress, agentic AI systems still lack principled, searchable capability profiles, making \emph{capability discovery} and partner selection ad hoc even in prominent orchestration frameworks~\cite{shen2023hugginggpt,wu2024autogen,wang2025internet}. \textit{Dynamic orchestration} remains partially solved: role-based systems improve structure yet often rely on manual wiring and do not couple decomposition with operational budgets~\cite{hong2024metagpt,li2023camel}. Meeting \emph{resource constraints} (latency, bandwidth, energy) alongside semantic fit is remarkable at the edge, where agents run on IoT devices~\cite{curasma2024agents}, motivating embedding-based \emph{semantic routing} and efficient, reliable transports~\cite{aws2025multillm,mqttv5,emqx2024harnessing,alotaibi2024secure}. Heterogeneous security and regulatory regimes further raise \emph{policy compliance} requirements, demanding auditable enforcement across agents and data boundaries~\cite{neupane2025towards}. At \emph{scale}, coordination overhead and loss of coherence emerge as agent counts and workflow depth grow, calling for sublinear retrieval and structure-aware coordination~\cite{xiong2025self,piao2025agentsociety,malkov2018efficient}. Operational \emph{observability and reliability} are also underdeveloped: real deployments report behavioural variability, drift, and governance gaps in agentic processes and ecosystems~\cite{fournier2025agentic,wong2025intelligent}. Finally, there are no interoperable protocols and ontologies to standardize capabilities and interactions, limiting portability across stacks and domains~\cite{kong2025survey,yang2025survey,hasan2025model,hou2025model}.

\paragraph{Our Approach.} FoA replaces static, topic-centric wiring with \emph{dynamic, capability-driven orchestration}, aligning with calls for semantics-first coordination in agent ecosystems~\cite{kong2025survey,yang2025survey,shen2023hugginggpt,wu2024autogen}. Agents publish (VCVs), structured, versioned profiles embedded in a high-dimensional space, so capabilities become searchable artifacts compatible with emerging interoperability efforts (e.g., Model Context Protocol (MCP)-based capability schemas)~\cite{hasan2025model,hou2025model}. We index VCVs using a sharded Hierarchical navigable small world (HNSW) index to support sublinear matching at scale while preserving nuanced distinctions among related skills~\cite{malkov2018efficient}. At dispatch time, FoA applies \emph{semantic routing} that couples profiles' similarities with policy checks and resource budgets (i.e., latency, bandwidth, energy consumption), rather than relying on keywords or static registries~\cite{aws2025multillm,neupane2025towards}. Operational feasibility is enforced with transport-aware choices for IoT settings, where the Message Queuing Telemetry Transport (MQTT) protocol provides efficient, reliable delivery under constrained networks~\cite{mqttv5,emqx2024harnessing,alotaibi2024secure}. For \emph{dynamic task decomposition}, FoA elicits candidate breakdowns from compatible agents and merges them into a consensual directed acyclic graph DAG, drawing on role-structured collaboration patterns from multi-agent systems~\cite{hong2024metagpt,li2023camel}. Finally, \emph{intelligent orchestration} optimizes assignments over semantic fit and operational cost, supporting centralized and decentralized modes~\cite{xiong2025self,piao2025agentsociety}; this aligns with distributed orchestration needs observed in infrastructures such as \textsc{CAFEIN}\textsuperscript{\textregistered}, CERN's federated AI platform, where privacy-preserving data access and cross-institution coordination are central constraints~\cite{cafein,santos2024federated}.

\paragraph{Contributions.} We make three technical contributions to enable capability-driven orchestration of heterogeneous agent federations: (1) \emph{Versioned Capability Vectors (VCVs)}, a machine-readable representation that transforms agent capabilities, costs, and constraints into searchable semantic embeddings indexed via sharded HNSW for sub-linear retrieval at scale. (2) \emph{Dynamic collaborative decomposition}, where compatible agents jointly propose subtask breakdowns that the orchestrator merges into a consensus DAG, coupling task structure with operational budgets rather than relying on static decomposition. (3) \emph{Smart clustering protocols} that group agents working on identical subtasks into collaborative channels for $k$-round refinement, balancing diverse perspectives against communication overhead through hierarchical similarity-based partitioning. Unlike prior works, which rely on manual wiring or role-centric dispatch without a searchable capability registry, FoA's VCVs enable policy and cost-aware \emph{semantic routing} and \emph{consensus DAG} execution; a registry and transport-level substrate for clustered collaboration with explicit policy labels. FoA achieves 13$\times$ improvement over single-model baselines on the HealthBench hard benchmark.

\section{Background and Related Works}\label{sec:related}

\paragraph{Agentic AI Systems: Communication Protocols and Architectures.} Agentic AI represents a paradigm shift from traditional AI systems that execute single tasks to systems that can plan, reason, and coordinate over extended horizons. Recent surveys differentiate \emph{AI agents}, modular systems optimized for specific tasks, from \emph{agentic AI systems}, which orchestrate collections of agents, decompose complex goals, and maintain persistent context across interactions~\cite{sapkota2025ai,schneider2025generative,gridach2025agentic}. The key insight is that while individual AI agents excel at specific tasks, complex problems often require the coordinated effort of multiple specialized agents~\cite{belcak2025small}. This motivates research on scalable coordination mechanisms, communication protocols, and orchestration strategies that can manage the complexity of multi-agent systems. A parallel research line examines communication protocols and architectures for LLM-driven agents. Both Yang et al.\cite{yang2025survey} and Kong et al.\cite{kong2025survey} provides a comprehensive analysis of agentic AI protocols, highlighting the lack of standardized communication methods between agents, tools, and data sources. Their classification distinguishes context-oriented protocols that support local tool invocation from inter-agent protocols that enable cooperative behaviour, and contrasts general-purpose versus domain-specific solutions. Comparative assessments reveal trade-offs between security, scalability, and latency across existing protocols and call for next-generation designs with adaptability, privacy preservation, and group-based interaction. Emerging work also demonstrates that successful agentic architectures require robust trust, risk, and security management frameworks~\cite{raza2025trism}, particularly as these systems increasingly operate in safety-critical domains. The theoretical foundations underlying agentic coordination suggest that these systems would benefit from established systems theory principles to manage their inherent complexity and emergent behaviours~\cite{miehling2025agentic}. Furthermore, practical implementations of agentic AI are showing promise across diverse application domains, from industrial automation~\cite{piccialli2025agentai} to healthcare compliance~\cite{neupane2025towards}, demonstrating the broad applicability of coordinated multi-agent approaches. Complementary work on  MCP~\cite{anthropic2024mcp,hou2025model,mcp2024transports,hasan2025model} proposes a standardized interface for connecting language models to external tools. However, most of the MCP implementations currently rely on HTTP and Server-Sent Events, introducing security concerns~\cite{narajala2025securing} and limiting their applicability to critical contexts or resource-constrained environments.

\paragraph{Multi-Agent Orchestration and Coordination} Researchers have explored various orchestration frameworks for coordinating agentic AI workflows across distributed systems. Multi-agent architectures leverage specialized agents with distinct capabilities: sensing, learning, reasoning, predicting, and executing that must be orchestrated to handle complex tasks through peer coordination~\cite{belcak2025small}. Recent frameworks like HuggingGPT~\cite{shen2023hugginggpt}, AutoGen~\cite{wu2024autogen}, and MetaGPT~\cite{hong2024metagpt} demonstrate sophisticated orchestration patterns where a central coordinator routes tasks to specialized agents based on their capabilities. Self-organizing agent networks~\cite{xiong2025self} enable dynamic workflow automation where agents autonomously form task-specific coalitions without centralized control. Communication protocols like CAMEL~\cite{li2023camel} promote structured inter-agent dialogue for collaborative problem-solving. Modern multi-LLM routing strategies employ various orchestration approaches, including static rule-based routing, dynamic model selection based on task complexity, and learned routing policies that optimize for performance and costs~\cite{aws2025multillm,li2025towards}. Despite their strengths, these frameworks hinge on fixed roles and hand-wired coordination, lacking a searchable capability registry, cost-/policy-aware semantic routing, and standardized collaboration protocols needed for enterprise-scale deployment.

\paragraph{MQTT as Transport Layer for Distributed Agent Systems} To provide a message passing layer between AI agents, the MQTT protocol~\cite{mqttv5} provides a stack of tools for large-scale distributed systems communication through its lightweight publish/subscribe semantics. Unified namespace (UNS) patterns leverage MQTT's topic hierarchy to create a semantic addressing scheme that eliminates data silos and enables seamless agent interoperability~\cite{peter2024impact}. Unlike traditional request-response architectures, MQTT's event-driven model naturally supports the asynchronous, many-to-many communication patterns required by multi-agent systems~\cite{emqx2024harnessing,wong2025intelligent}. Recent developments extend MCP over MQTT, replacing HTTP's synchronous constraints with MQTT's scalable pub/sub architecture for distributed agent coordination~\cite{emqx2025mcp-mqtt,ji2025emqx-mcp-server,emqx2025mcp-gateway,espressif2025esp32mcp}. This transport choice enables capability discovery across thousands of heterogeneous agents while maintaining sub-second latency and reliable message delivery under varying network conditions~\cite{alotaibi2024secure,curasma2024agents,elewah2025agentic}. The protocol's inherent support for Quality of Service guarantees, retained messages, and wildcard subscriptions makes it particularly suited for orchestrating global-scale agent federations~\cite{kong2025survey,yang2025survey,hasan2025model,hou2025model}.

\begin{figure}[t]
    \centering
    \vspace{-10pt}
    \includegraphics[width=.9\linewidth]{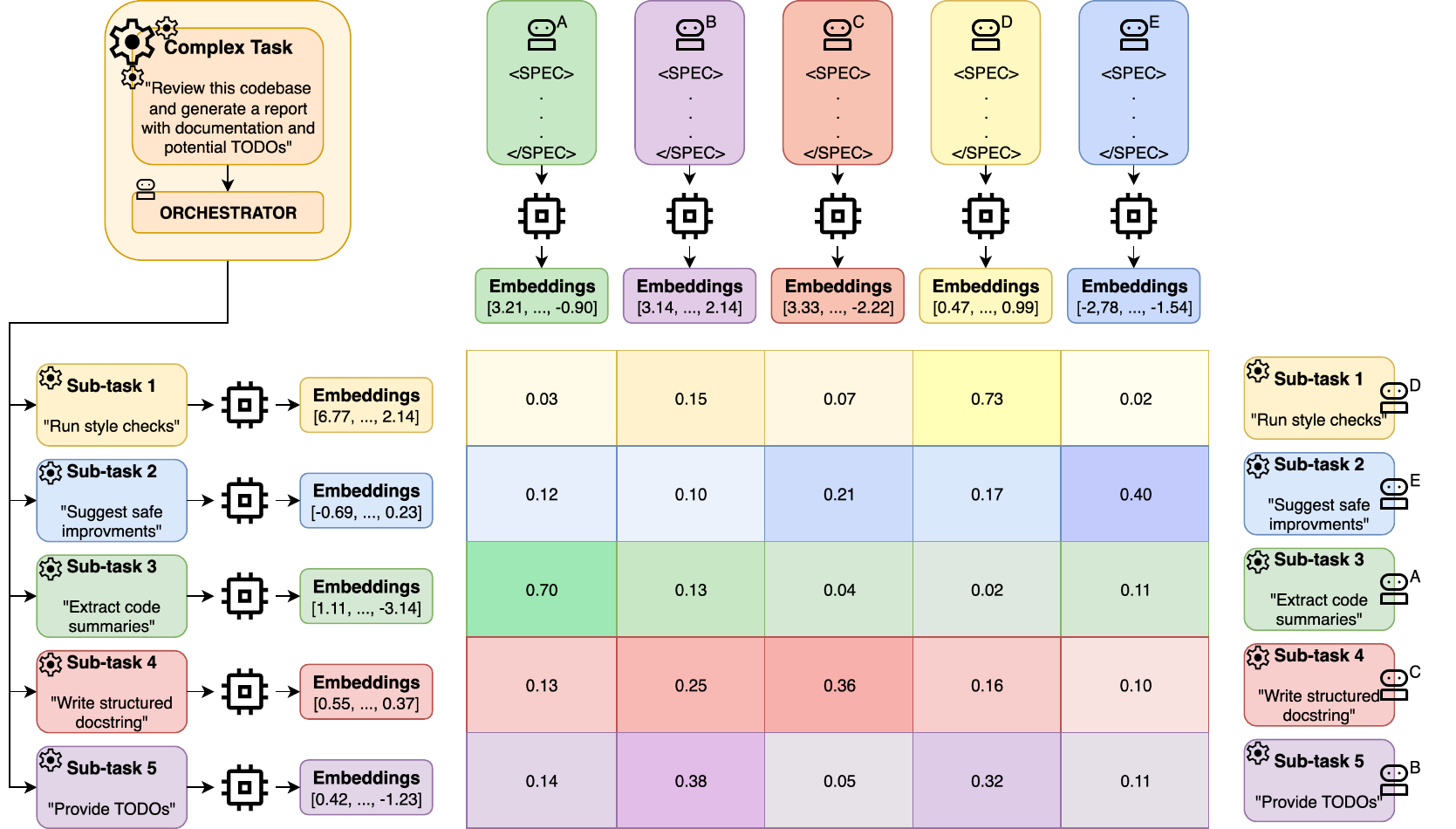}
    \caption{\textbf{Orchestrator-driven Sub-task Decomposition and Semantic Routing.} The orchestrator receives a complex task, proposes subtasks (1-5), embeds each subtask, and compares them to agent capability embeddings to form a similarity matrix. Using similarity, cost, and policy constraints, it assigns each subtask to the most compatible agent or cluster (right-hand mapping).}
    \label{fig:orch}
    \vspace{-10pt}
\end{figure}

\section{Federation of Agents System Architecture}
This section details the core artifacts and mechanisms that compose our semantics-aware orchestration. We first introduce the artifacts' Specs that guide each agent's behaviour, then describe how these are embedded into our searchable profiles' representation. Furthermore, we characterise the orchestrator (Agent-0) and the worker agents (Agent-1) before discussing the formation of semantic clusters and the DAG that governs execution order (Fig.~\ref{fig:foe}). In this work, \emph{federation of agents} refers to runtime orchestration across heterogeneous agents and transports with policy-guarded data flow and capability discovery. This is distinct from \emph{federated learning} (FL), which concerns privacy-preserving model training. FoA can optionally include an FL \emph{training} phase by storing model-update and differential-privacy parameters in VCVs and scheduling training/aggregation as DAG nodes; our experiments in this paper focus on the \emph{inference-time} orchestration setting.

\paragraph{Artifacts.} Each agent is associated with a \emph{model specification}, or \emph{Spec}, a document listing the agent's goals, tools, resources, rules, and principles. Inspired by alignment frameworks emphasising that language assistants should be \emph{helpful}, \emph{harmless} and \emph{honest}~\cite{ouyang2022training,askell2021general}, a Spec typically combines high-level imperatives ("assist the user"; "do not break the policies") with detailed dos and don'ts (e.g., avoid certain words, report emergencies, respect privacy). Recent work on reinforcement learning from human feedback (RLHF) shows that language models can be fine-tuned to follow such instructions by leveraging human preference data and learning a reward model~\cite{ouyang2022training}; the resulting models more often adhere to the helpful, honest, harmless (HHH) criteria than models trained purely on next-token prediction~\cite{askell2021general}. In our framework, each agent memorises its Spec through such alignment techniques and learns to reason over it to follow instructions, refuse harmful requests, and resist the temptation to hallucinate citations or fabricate task completion (honest) as emphasised in alignment studies~\cite{ouyang2022training,askell2021general,yang2024alignment}. Specifications become machine-readable by embedding them into our VCVs using a specialised tokenizer and passed through a sentence-embedding model~\cite{nussbaum2024nomic,choi2025embeddinggemma} to obtain a dense vector summarising the agent's profile. This \emph{spec embedding} augments the agent's skill and resource representations, ensuring that subsequent semantic routing takes account of both functional ability and behavioural constraints. Embedding Specs allows the orchestrator to query not only \emph{"who knows how?"} but also \emph{"who aligns with what policies?"}.

\paragraph{Versioned Capability Vectors.}  A Versioned Capability Vector encodes the state of an agent in a searchable form. For agent $a_i$, we define
\begin{equation}
  \mathrm{VCV}_{a_i} = \bigl(\mathbf{c}_{a_i}, \mathbf{s}_{a_i}, \mathbf{r}_{a_i}, \mathbf{p}_{a_i}, \mathbf{e}_{a_i}, v_{a_i}\bigr),  
\end{equation}
  
where $\mathbf{c}_{a_i} \in \mathbb{R}^d$ is a dense capability embedding describing the agent's core competencies, $\mathbf{s}_{a_i} \in \{0,1\}^{\ell}$ is a Bloom filter over discrete skills, $\mathbf{r}_{a_i} \in \mathbb{R}^m$ records resource requirements and quality-of-service guarantees (e.g., latency budget, energy consumption), $\mathbf{p}_{a_i} \in \{0,1\}^p$ encodes policy compliance flags (e.g.\ regulatory and security labels), $\mathbf{e}_{a_i} \in \mathbb{R}^{d'}$ is the spec embedding described above, and $v_{a_i} \in \mathbb{N}$ is a version counter. The capability embedding $\mathbf{c}_{a_i}$ is produced by feeding a natural-language description of the agent's abilities through a pre-trained language model $\psi: \mathcal{G}\to \mathbb{R}^d$. Each time an agent updates its capabilities or specification, its version number increments, and a new VCV is broadcast.

\paragraph{Orchestrator (Agent-0).} We call the orchestrator of the federation Agent-0 (A-0). When an external task arrives, the orchestrator consults the current VCV index to discover candidate agents, performs dynamic decomposition, forms clusters, and orchestrates their collaboration. It maintains a sharded HNSW index over VCV embeddings to support sub-linear retrieval at scale. It also runs a lightweight \emph{$\Delta$-gossip} protocol to propagate VCV updates: rather than broadcasting full capability vectors, agents periodically exchange only the deltas between their local VCV sets. Task metadata and VCV updates are disseminated over MQTT \texttt{foa/meta} and \texttt{foa/retain} topics, respectively, allowing asynchronous, broadcast communications. During execution, the orchestrator subscribes to cluster channels (described in Sec.~\ref{sec:flow}) and collects completion signals and intermediate artefacts. It merges partial results along a sub-task execution graph by traversing it in topological order, ensuring that dependencies are satisfied before downstream subtasks are synthesised. At completion, it publishes the final output on a \texttt{foa/result} topic, updates individual agents' reputations based on grading feedback before disposing of them.

\paragraph{Decomposition and scheduling.} The orchestration process begins when Agent-0 receives a task $t$ from the environment. Rather than attempting to solve complex tasks monolithically, our system employs a hierarchical decomposition strategy that leverages the collective intelligence of the agent pool. Agent-0 first embeds the task description into a semantic space, then initiates a cooperative planning routine where multiple specialized agents propose candidate decompositions based on their domain expertise. These perspectives are synthesized into a consensual directed acyclic graph $G = (V, E)$, where vertices represent atomic subtasks and edges encode execution dependencies. When the task structure is established, the orchestrator performs optimal agent assignment. To capture unique capabilities, resource constraints, and performance characteristics, we introduce a scoring function that evaluates agent-subtask pairs across four complementary dimensions:
\begin{equation}
\alpha_{s_i,a_j} = \operatorname{sim}\bigl(\mathbf{c}_{s_i},\mathbf{c}_{a_j}\bigr)\, \cdot\, \mathbb{I}\bigl[\mathbf{p}_{s_i} \subseteq \mathbf{p}_{a_j}\bigr]\, \cdot\, f\bigl(\mathbf{r}_{s_i}, \mathbf{r}_{a_j}\bigr)\,\cdot\, g\bigl(\mathbf{e}_{s_i}, \mathbf{e}_{a_j}\bigr),
\end{equation}
where $\operatorname{sim}$ measures semantic alignment between the subtask requirements and agent capabilities in the embedding space, the indicator $\mathbb{I}$ ensures policy compliance (required permissions $\mathbf{p}_{s_i}$ are within the agent's authorization set $\mathbf{p}_{a_j}$), \emph{$f$ is instantiated as a monotone penalty over resource gaps} (latency, bandwidth, memory), decreasing as $\|\mathbf{r}_{s_i}-\mathbf{r}_{a_j}\|$ increases, and \emph{$g$ is instantiated as cosine similarity} between specification embeddings $g(\mathbf{e}_{s_i},\mathbf{e}_{a_j})=\cos(\mathbf{e}_{s_i},\mathbf{e}_{a_j})\in[-1,1]$ after $\ell_2$-normalization. This makes the gate explicit: any failed policy check zeros the score; otherwise, the score is down-weighted by resource mismatches and boosted by spec alignment. The orchestrator then transforms these compatibility scores into a concrete execution plan through constrained optimization. Formally, given $k = |\mathcal{S}|$ subtasks and $n = |\mathcal{A}|$ available agents, Agent-0 computes an optimal binary assignment matrix $\mathbf{X} \in \{0,1\}^{k \times n}$ that maximizes the expected utility:
\begin{equation}
\max_{\mathbf{X}} \sum_{i=1}^k \sum_{j=1}^n x_{ij}\, \alpha_{s_i,a_j} \,\mathrm{perf}_{a_j},
\end{equation}
subject to capacity constraints and \emph{at-most-$r_i$} team size per subtask: $\sum_{j} x_{ij} \in [1, r_i]$ for all $i$ (single-agent when $r_i{=}1$). The resulting $\mathbf{X}$ gives the candidate team for each $s_i$; \emph{cluster formation} (Sec.~\ref{sec:flow}) then groups selected agents solving the same $s_i$ for $k$-round collaborative refinement before synthesis. This resolves the earlier single-agent/cluster tension by making teams first-class and clusters a subsequent refinement stage.

\paragraph{Agents (Agent-1).} Agents form the workforce of the federation. Each Agent-1 (A-1) is wrapped around a pre-aligned language model using Group Relative Policy Optimization (GRPO)~\cite{shao2024deepseekmath}. The alignment leverages its Spec as a reward function and local domain data (documents, databases, and API outputs) to specialise the agent's behaviour. Agents have access to a tool-use controller that automatically selects retrieval or computation tools (e.g., vector search, SQL query, web API) based on the current task; the available tools and corpora are reflected in the agent's VCV resource vector $\mathbf{r}_{a_i}$. Internally, each agent maintains a scratchpad to store intermediate reasoning and implements the instructions in its Spec; for example, it declines unsafe requests and signals uncertainty when the Spec mandates honesty. When assigned a subtask, Agent-1 generates an initial draft solution by integrating its model output with the retrieved context. It then participates in cluster-based refinement rounds. To encourage diverse perspectives and reduce hallucinations, we group agents solving the same subtask into semantic clusters. Given a set of candidate agents $\mathcal{A}_{s_i}$ for subtask $s_i$, we compute a similarity matrix combining (i) the cosine similarity between their capability embeddings $\mathbf{c}_{a_j}$, (ii) the similarity of their preliminary outputs (once available) and (iii) the overlap of their spec embeddings $\mathbf{e}_{a_j}$. Hierarchical clustering on this matrix yields clusters $C_1, \dots, C_m$ of A-1 agents. Within each cluster, agents exchange intermediate drafts and critiques for $k$ rounds using the topic \texttt{foa/clusters/\{cluster\_id\}/channel}. After each round, they vote on whether to stop; once consensus is reached, the cluster emits a \texttt{TASK\_COMPLETE} signal and returns a refined subtask result. Clustering thus enables collaborative refinement akin to peer review while limiting communication overhead.

\section{Federation of Agents Execution Flow}\label{sec:flow}

In this section, we analyze the life cycle of a single task handled by FoA. The framework orchestrates the end-to-end execution of a complex problem through a six-phase pipeline that captures decomposition, drafting, collaboration, and synthesis. Formally, given an incoming task $t$ provided by the environment and a set of agents $\mathcal{A}$ equipped with VCV, the orchestrator A-0 establishes a DAG $G = (\mathcal{S},E)$ of sub-task execution order whose vertices correspond to (sub-task, A-1) pairs and whose edges encode relational dependencies between sub-tasks. Each phase described below operates on $G$ and updates its state until all nodes get detached from $G$ by completing sub-tasks or by receiving a \texttt{DISPATCH} signal.

\paragraph{Sub-task decomposition, consensus, and assignment.} Upon receiving a task $t$ from the environment, Agent-0 embeds its natural-language description into the semantic space and queries the VCV index for candidate agents. Each compatible agent $a_j$ returns a proposal consisting of a set of subtasks $\mathcal{S}_{a_j}$ and a set of dependencies $E_{a_j}$ describing how those subtasks should be ordered. Agent-0 collects these proposals, merges them via a consensus mechanism, and validates acyclicity, thereby producing a global DAG $G=(\mathcal{S},E)$ with $\mathcal{S} = \bigcup_j \mathcal{S}_{a_j}$ and $E = \bigcup_j E_{a_j}$. The orchestrator then solves the assignment problem introduced in Sec.~\ref{sec:intro}: it computes scores $\alpha_{s_i,a_j}$ for each subtask-agent pair based on semantic alignment, policy compliance, resource fit, and specification similarity. Solving the resulting integer program yields an assignment matrix $\mathbf{X} \in \{0,1\}^{|\mathcal{S}| \times |\mathcal{A}|}$ that maps each subtask $s_i$ to A-1 agents while respecting capacity constraints. FoA organises execution around $G$: leaves (subtasks with no incoming edges) can begin immediately, whereas internal nodes $s_i$ wait until all predecessors $s_j$ with $(s_j\to s_i)\in E$ have reported completion. A \texttt{SYNTH} tool combines results from predecessors when triggering a downstream subtask, enabling partial results to propagate forward without blocking unrelated branches and facilitating fine-grained concurrency in large workflows.

\paragraph{First draft with resource access.} Once assigned to a subtask $s_i$, each A-1 retrieves relevant context from its local resources, such as databases or external tools, via tool-use controllers embedded in A-1. Conditioning on this context and its specification embedding $\mathbf{e}_{a_j}$, the agent produces a first-draft answer $d_{i}^{(j)}$ for the subtask. This draft anchors subsequent refinement, which is posted to a cluster-specific MQTT channel associated with $s_i$. The retrieval step could implement an additional resource-aware step: the agent consults its resource vector $\mathbf{r}_{a_j}$ to adjust re-spawn parameters, ensuring that within clusters, A-1 size (in GB of GPU vRAM), throughput velocity (in tok/sec), context window of the channel, and maximum budget of available tokens to produce solutions do not exceed latency or energy constraints. As shown later in Sec.~\ref{sec:experiments}, we found it practical to set-up A-1 as a pre-aligned small language model ($\leq20B$ params) for fast-feedback execution of multiple refinement rounds over long-term horizons~\cite{ouyang2022training, ji2023ai, belcak2025small}.

\begin{figure}[t]
    \centering
    \vspace{-10pt}
    \includegraphics[width=.8\linewidth]{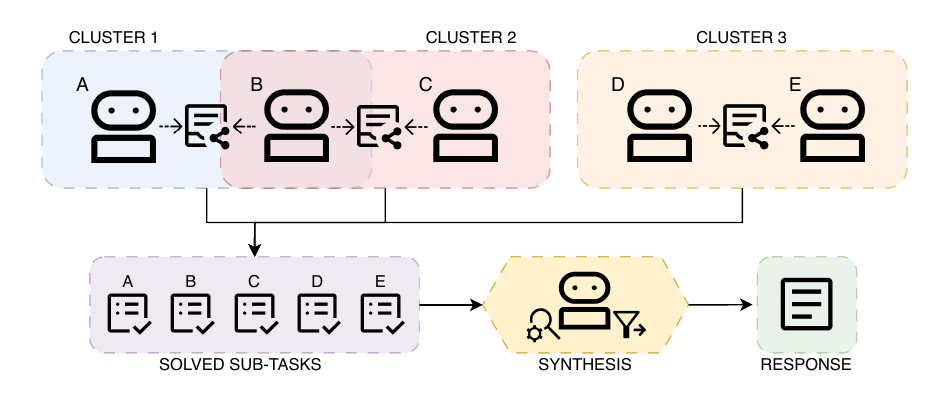}
    \caption{\textbf{Collaborative refinement inside a high-similarity cluster.} Agents with closely aligned capabilities on the same subtask are grouped into a cluster, exchange drafts and critiques for $k$ refinement rounds, and upon consensus emit \texttt{TASK\_COMPLETE} to return a final subtask result to the orchestrator for synthesis.}
    \label{fig:clust}
    \vspace{-10pt}
\end{figure}

\paragraph{Cluster formation via semantic similarity.} Agent-0 groups the agents assigned to the same subtask into collaborative clusters based on their capability vectors and preliminary outputs. Concretely, for subtask $s_i$ with assigned agents $\mathcal{A}_{s_i}$, we compute a similarity matrix combining (i) the cosine similarity of their capability embeddings $\mathbf{c}_{a_j}$, (ii) the cosine similarity of their draft embeddings and (iii) the overlap of their spec embeddings $\mathbf{e}_{a_j}$. Hierarchical clustering on this matrix yields clusters $C_1,\dots,C_m$ of size chosen to balance diversity against coordination overhead. A dedicated cluster channel \texttt{foa/clusters/\{cluster\_id\}/channel} is created on the MQTT broker for each cluster, allowing members to share messages without interfering with other topics. Fig.~\ref{fig:clust} illustrates the refinement process inside a high-similarity cluster.

\paragraph{Intra-cluster execution} Within each cluster $C_j$, agents iteratively refine their drafts. At round $r$, every A-1 agent posts its current draft to the cluster channel and receives the drafts of its peers (Fig.~\ref{fig:orch}). Agents critique and update their own answers by integrating insights from others, using simple majority voting or reputation-weighted aggregation to decide which components to adopt. Formally, let $\mathcal{M}_{s_i,C_j}^{(r)}$ denote the multiset of messages exchanged at round $r$; refinement continues for $k$ rounds or until a consensus signal is triggered. Throughout this process, agents adhere to their Specs: they refuse to produce unsafe content, annotate uncertainties when appropriate, and propagate provenance metadata with each message. This collaborative refinement functions as a peer-review cycle that aims to improve factuality and reduce hallucinations.

\paragraph{Reporting to the orchestrator.} When the agents in cluster $C_j$ reach consensus on a refined answer $\hat{d}_{i}^{(j)}$ for subtask $s_i$, they emit a \texttt{TASK\_COMPLETE} message on their cluster channel. This message contains the final answer along with evaluation metrics (e.g., confidence scores) computed by the cluster. Agent-0 subscribes to all cluster channels and listens for these completion signals. Upon receiving \texttt{TASK\_COMPLETE} for $s_i$, it marks the node as finished in the DAG $G$ and stores the result for use by downstream subtasks. If no consensus is reached within a predefined timeout, A-0 either reassigns the subtask to another agent or accepts the highest-scoring draft according to its evaluation function, thereby preventing deadlock.

\paragraph{Result synthesis.} After all clusters have reported completion, A-0 traverses $G$ in topological order. For each subtask $s_i$, it invokes the \texttt{SYNTH} operator to combine the results of its predecessor subtasks with the refined answer for $s_i$:
\begin{equation}
\mathbf{sol}_{s_i} = \texttt{SYNTH}\bigl(\{\mathbf{sol}_{s_j} : (s_j\to s_i) \in E\}\cup \{\hat{d}_{i}^{(j)}\}\bigr).
\end{equation}

This operator may concatenate texts (default solution), resolve conflicting assertions across cluster outputs (i.e., \texttt{rebase}), or summarise divergent perspectives into a unified answer (i.e., \texttt{merge}). We implement \texttt{SYNTH} via meta-prompting~\cite{suzgun2024meta} by steering the internal chain-of-thought of A-0~\cite{wei2022chain}. Once all leaves in the DAG are executed and their results propagated forward, the final answer for the original task $t$ is obtained at the root of $G$. A-0 then publishes the final solution on \texttt{foa/result} MQTT topic, updates the reputations of participating agents based on the quality of their contributions, and archives their updated VCVs for future routing decisions. Many phases are parallelized and cluster sizes are capped (3-5), supporting horizontal scalability before large-scale deployment. We quantify end-to-end complexity in Appendix~(\ref{sec:algos}). 

\section{Evaluation and Experimental Results}\label{sec:experiments}

We evaluate the FoA framework on OpenAI's HealthBench Hard~\cite{arora2025healthbench}, a comprehensive benchmark for assessing language models in healthcare contexts. HealthBench Hard comprises 1,000 multi-turn conversations between models and users (both healthcare professionals and patients), with responses evaluated against physician-written rubrics spanning 48,562 unique criteria. The rubrics assign positive or negative points depending on whether a response satisfies desirable or undesirable criteria; scores range between $-10$ and $10$ and are combined into a per-example score by a model-based grader that has been validated against physician judgments. The overall HealthBench score is obtained by averaging per-example scores and clipping the mean to the range $[0,1]$. Unlike traditional multiple-choice medical benchmarks, HealthBench contains an open-ended nature of real healthcare interactions through conversation-specific evaluation across seven themes (emergency referrals, context seeking, global health, health data tasks, expertise-tailored communication, responding under uncertainty, and response depth) and five behavioural axes (accuracy, completeness, context awareness, communication quality, and instruction following). In Sec.~\ref{sec:add-exp}, we provide additional details on specific configuration parameters and implementation decisions. 

\begin{figure}[t]
    \centering
    \includegraphics[width=.8
    \linewidth]{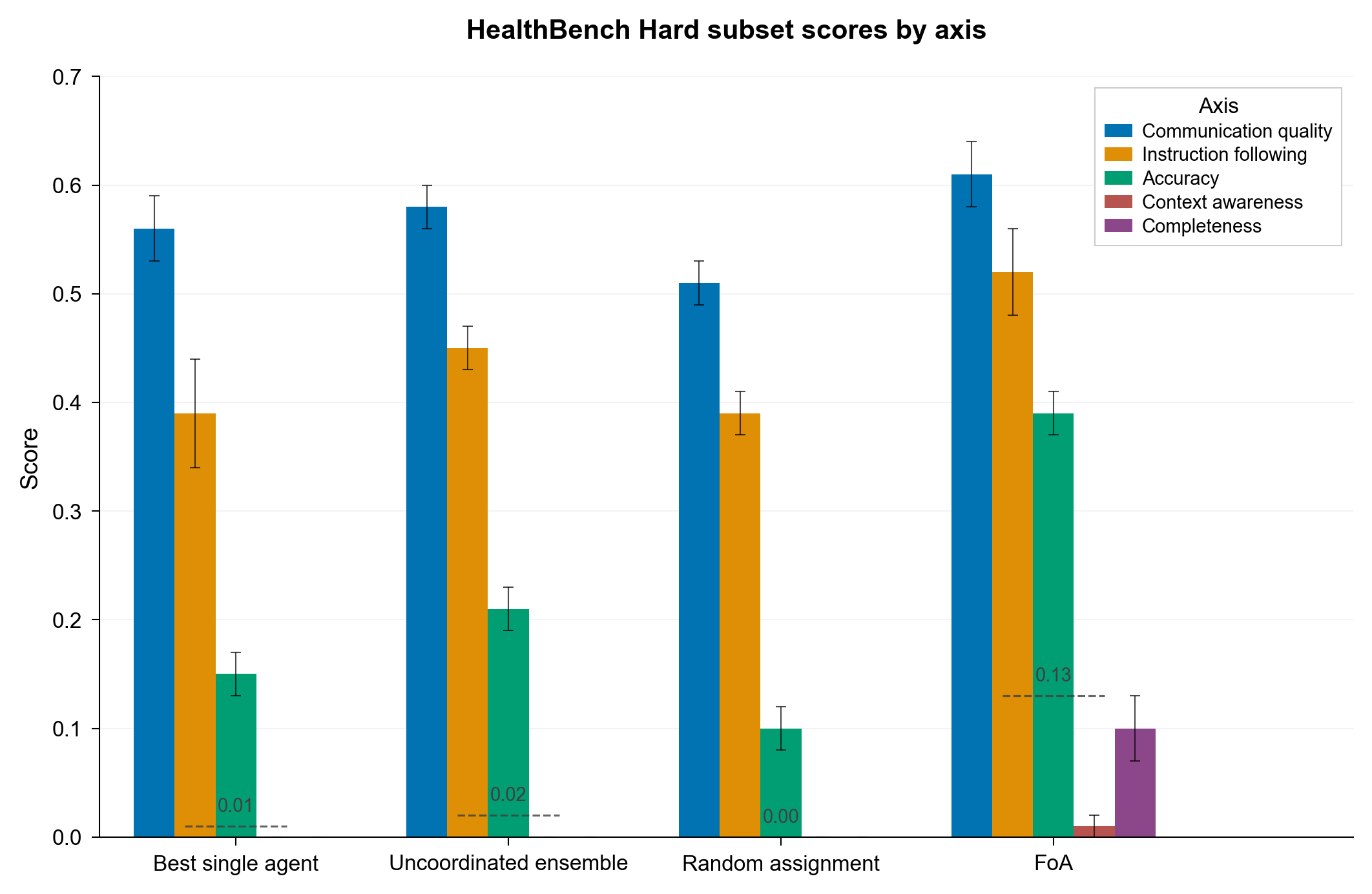}
    \caption{\textbf{HealthBench Hard results.} We report the overall score (mean~$\pm$~bootstrap s.d.) and per-axis scores for FoA and baselines. Individual example scores can be negative, but average scores are clipped to zero.}
    \label{fig:main-results}
    \vspace{-5pt}
\end{figure}

\paragraph{Main Results.} Fig.~\ref{fig:main-results} summarises the performance of FoA and the baselines on HealthBench Hard. FoA achieves an overall score of $0.13$, a \emph{13x} relative improvement over the best single agent baseline (Medgemma~\cite{sellergren2025medgemma}) and a \emph{6.5x} improvement over the uncoordinated ensemble. Random assignment performs markedly worse, underscoring the importance of capability-aware routing. FoA consistently outperforms baselines across all seven themes. The collaborative cluster protocol particularly benefits high-stakes questions where multiple perspectives improve accuracy and context awareness.

\section{Conclusion \& Discussion}\label{sec:conclusion}
We presented Federation of Agents (FoA), a semantics-aware communication fabric that enables dynamic, capability-driven orchestration of large-scale multi-agent AI systems. Through machine-readable Versioned Capability Vectors (VCVs) and cost-aware semantic routing, FoA transforms static topic-based coordination into a scalable infrastructure for efficient reasoning over extended horizons, solving a coordination problem in agentic AI regarding \emph{who can do what, at what cost, and with what reputation?}

\paragraph{Limitations and Open Research Questions.} Several challenges limit our current approach and represent important directions for the community. The effectiveness of semantic routing is bounded by embedding quality, creating a cold-start problem where agents with novel capabilities may remain underutilized until sufficient interaction data is collected. Our clustering algorithm may form sub-optimal groups when task similarity metrics fail to capture true collaboration potential, triggering a fallback mechanism that often leads to reduced effectiveness in heterogeneous domains. The VCV representation, while expressive, may not capture complex compositional capabilities or dynamic skill emergence during agent execution. Communication overhead scales quadratically within clusters, limiting the practical cluster size to 3-5 agents. Additionally, while our reputation and policy enforcement mechanisms provide robust security against honest-but-curious adversaries, sophisticated attacks such as coordinated Sybil networks or adversarial capability misrepresentation remain open challenges that can be mitigated by using sandbox executions~\cite{santos2024federated}.

\paragraph{Future Research Directions.} Our immediate research focuses on four critical frontiers. First, we will develop reinforcement learning-based adaptive routing controllers that can dynamically optimize cost-biasing, matching thresholds, and clustering parameters based on real-time network performance and evolving task priorities. Second, we plan to explore cross-cluster communication protocols to enable knowledge sharing between related clusters while maintaining scalability. Third, we will integrate zero-knowledge proof systems and trusted execution environments to enable verifiable capability attestations, directly addressing the trust assumptions in our current design, integrating the security and safety measures of~\cite{santos2024federated} in FoA.

\paragraph{Broader Impact and Societal Considerations.} Large-scale federations of AI agents present both transformative opportunities and significant risks. On the positive side, democratizing access to specialized AI capabilities could accelerate scientific discovery, improve resource allocation efficiency, and enable more responsive technological infrastructures. However, the concentration of autonomous decision-making capabilities raises concerns about algorithmic bias amplification, privacy preservation across agent boundaries, and the increasing opacity of accountability structures. The governance mechanisms embedded in FoA, including auditable provenance trails and policy-as-code enforcement, represent necessary but insufficient steps toward responsible deployment. The broader AI community must engage in continued dialogue about appropriate legal frameworks, ethical guidelines, and oversight mechanisms for autonomous agentic ecosystems.

In conclusion, the Federation of Agents provides a theoretically grounded and practically scalable approach to multi-agent coordination that transforms capabilities and constraints into a searchable, auditable substrate. Our smart clustering innovations demonstrate that collaborative refinement can significantly improve solution quality while maintaining computational tractability. Addressing core challenges in semantic routing, distributed orchestration, intra-agent collaboration, and trust management establishes a foundation for the next generation of collaborative AI systems. The 10x performance improvements on HealthBench with respect to the best single agent validate the practical benefits of capability-driven orchestration with smart clustering. We invite the research community to build upon these contributions as we collectively advance toward more capable, trustworthy, and socially beneficial agentic AI ecosystems.

\newpage
\bibliographystyle{unsrt}
\bibliography{refs}

\newpage

\appendix

\section{Detailed Algorithm Analysis}\label{sec:algos}

In this section, we formalise each stage of the FoA execution pipeline from an algorithmic perspective. Our analysis clarifies the internal mechanics and quantifies the computational complexity to link the design to prior literature. We follow the six phases outlined in Section~\ref{sec:flow}: (1) dynamic task decomposition, (2) first draft generation, (3) cluster formation, (4) intra-cluster consensus, (5) reporting, and (6) synthesis.

\paragraph{Phase~1: Dynamic task decomposition and consensus.}
Given an input task $\texttt{t}$, the orchestrator searches for compatible agents and constructs a directed acyclic graph (DAG) of subtasks. The procedure, formalised in Algorithm~\ref{alg:task_decomposition}, embeds the task into the capability space, evaluates each agent's compatibility score against semantic, policy, and resource criteria, and queries a sharded HNSW index to retrieve the most promising candidates. Selected agents propose candidate subtask sets and dependency structures; these proposals are merged via a consensus mechanism into a single DAG and validated for acyclicity and consistency. Our design draws inspiration from recent multi-agent frameworks that dynamically decompose complex tasks into manageable subtasks and tailor subagents for each subproblem. With this dynamic decomposition based on intermediate results and dedicated subagents assignments, FoA avoids the error propagation and static limitations of preliminary decomposition schemes.

\begin{algorithm}[!ht]
\centering
\caption{Dynamic Task Decomposition Protocol}
\label{alg:task_decomposition}
\small
\begin{algorithmic}[1]
\Require Task $\texttt{t} \in \mathcal{T}$,  Set of A-1 Agents $\mathcal{A}$, threshold $\tau$, policy constraints $\mathbf{P}$
\Ensure Task DAG $G = (V,E)$ with subtasks $V=\mathcal{S}$ and dependencies $E$
\State \textbf{Phase~1a:} Task embedding and agent scoring
\State $\mathbf{c}_\texttt{t} \gets \psi(\texttt{t})$ \Comment{embed task description into capability space}
\State $\mathcal{A}_{comp} \gets \emptyset$ \Comment{candidate agent set}
\For{each $a_i \in \mathcal{A}$}
  \State $\alpha_{t,a_i} \gets \operatorname{sim}(\mathbf{c}_t, \mathbf{c}_{a_i}) \cdot \mu(\mathbf{P}, \mathbf{p}_{a_i}) \cdot \text{cost}(\mathbf{r}_t, \mathbf{r}_{a_i})$
  \If{$\alpha_{t,a_i} > \tau$}
    \State $\mathcal{A}_{comp} \gets \mathcal{A}_{comp} \cup \{a_i\}$ \Comment{retain compatible agents}
  \EndIf
\EndFor
\If{$|\mathcal{A}_{comp}|=0$}
  \State $\mathcal{A}_{comp} \gets \text{TopK}(\mathcal{A}, \operatorname{sim}(\mathbf{c}_t,\mathbf{c}_{a_j}), k)$ \Comment{fallback retrieval}
\EndIf

\State \textbf{Phase~1b:} Collaborative proposal and merge
\State $\mathcal{S}_{prop} \gets \emptyset$, $E_{prop} \gets \emptyset$
\For{each $a_i \in \mathcal{A}_{comp}$}
  \State $\text{subtasks}_{a_i} \gets \text{DECOMPOSE}(t, \mathbf{c}_{a_i})$ \Comment{agent-specific decomposition}
  \State $\text{deps}_{a_i} \gets \text{ANALYZE\_DEPS}(\text{subtasks}_{a_i})$ \Comment{local dependency analysis}
  \State $\mathcal{S}_{prop} \gets \mathcal{S}_{prop} \cup \text{subtasks}_{a_i}$
  \State $E_{prop} \gets E_{prop} \cup \text{deps}_{a_i}$
\EndFor
\State $G \gets \text{SYNTH\_PROPOSALS}(\mathcal{S}_{prop}, E_{prop})$ \Comment{consensus on subtask set and edges}
\State $G \gets \text{VALIDATE\_DAG}(G)$ \Comment{remove cycles and inconsistencies}
\State \Return $G$
\end{algorithmic}
\end{algorithm}

The time complexity of Phase~1 is dominated by evaluating compatibility
scores and merging proposals. Let $n=|\mathcal{A}|$ and $m$ be the
number of candidate agents retrieved; computing similarities for all
agents requires $O(n\,d)$ operations where $d$ is the embedding
dimension. Each agent proposes $O(|\mathcal{S}|)$ subtasks; merging
proposals requires at most $O(m\,|\mathcal{S}|\log |\mathcal{S}|)$ time
to reconcile duplicates and validate the DAG. Dynamic decomposition
improves robustness by allowing later phases to adjust subtasks based on
intermediate outputs, a key advantage over static splitting.

\paragraph{Phase~2: First draft generation with resource access.} Once the DAG is established and subtasks are assigned to agents, each agent retrieves context and produces an initial answer. The process resembles iterative reasoning methods such as ReAct~\cite{yao2023react}, Reflexion~\cite{shinn2023reflexion}, or Graph of Thoughts~\cite{besta2024graph}, where intermediate thoughts guide subsequent actions. However, in FoA, the first draft is produced individually by each assigned agent before collaborative refinement. Algorithm~\ref{alg:first_draft} formalises this stage.

\begin{algorithm}[!htb]
\centering
\caption{First Draft Generation}
\label{alg:first_draft}
\small
\begin{algorithmic}[1]
\Require Subtask $s \in \mathcal{S}$, assigned agent set $\mathcal{A}_s$
\Ensure Draft set $\mathcal{D}_s$ containing one draft per agent
\State $\mathcal{D}_s \gets \emptyset$
\For{each agent $a_i \in \mathcal{A}_s$}
  \State $\text{context} \gets \text{RETRIEVE\_RESOURCES}(s, a_i)$ \Comment{query local resources, use tools}
  \State $\text{draft} \gets \text{GENERATE\_ANSWER}(s, \text{context}, \text{Spec}_{a_i})$ \Comment{LLM inference conditioned on Spec}
  \State $\mathcal{D}_s \gets \mathcal{D}_s \cup \{(a_i, \text{draft})\}$
\EndFor
\State \Return $\mathcal{D}_s$
\end{algorithmic}
\end{algorithm}

Each agent's retrieval call and model inference contributes to the complexity of this phase. If $r_i$ denotes the retrieval cost and $t_i$ denotes the model run time for agent $a_i$, then the total complexity is $O(\sum_{a_i \in \mathcal{A}_s}(r_i + t_i))$. Because draft generation is parallel across agents, it scales horizontally with the number of assigned agents.

\paragraph{Phase~3: Cluster formation via semantic similarity.} To encourage team flow while controlling communication overhead, FoA groups agents working on the same subtask into clusters based on similarity across multiple dimensions (capability, resource cost, initial draft quality, and specification)~\cite{van2022promoting}. The orchestrator builds a similarity matrix for $\mathcal{A}_s$ and performs hierarchical clustering to partition agents into clusters. Algorithm~\ref{alg:cluster} formalises this procedure.

\begin{algorithm}[!ht]
\centering
\caption{Semantic Cluster Formation}
\label{alg:cluster}
\small
\begin{algorithmic}[1]
\Require Subtask $s$, agents $\mathcal{A}_s$, draft set $\mathcal{D}_s$, weight vector $(w_1,w_2,w_3,w_4)$
\Ensure Cluster partition $\mathcal{C}_s = \{C_1,\dots,C_m\}$
\State $n \gets |\mathcal{A}_s|$
\State Initialize $S \in \mathbb{R}^{n\times n}$ \Comment{similarity matrix}
\For{each pair $i < j$}
  \State $S_{i,j} \gets w_1 \cdot \cos(\mathbf{c}_{a_i}, \mathbf{c}_{a_j}) + w_2 \cdot \cos(\mathbf{r}_{a_i}, \mathbf{r}_{a_j})$
  \State $\phantom{S_{i,j} \gets}\ + w_3 \cdot \cos(\psi(\text{draft}_{a_i}), \psi(\text{draft}_{a_j})) + w_4 \cdot \cos(\mathbf{e}_{a_i}, \mathbf{e}_{a_j})$
  \State $S_{j,i} \gets S_{i,j}$
\EndFor
\State $\mathcal{C}_s \gets \text{HIER\_CLUSTER}(S)$ \Comment{agglomerative clustering with cut threshold}
\State \Return $\mathcal{C}_s$
\end{algorithmic}
\end{algorithm}

Constructing the similarity matrix requires $O(n^2 d)$ operations, and agglomerative clustering adds an $O(n^2 \log n)$ factor, yielding a total complexity of $O(n^2(d + \log n))$, consistent with the analysis in the main text. 

\paragraph{Phase~4: Intra-cluster execution and consensus.}
Within each cluster, agents iteratively share and refine their solutions until consensus is reached. This process can be viewed as a distributed consensus protocol in which agents negotiate a common answer through repeated information exchange. Such protocols have deep roots in distributed control theory: consensus algorithms guarantee that agents converge to a shared value through local information exchange under suitable connectivity assumptions. Algorithm~\ref{alg:intra_cluster} codifies our intra-cluster execution.

\begin{algorithm}[!htb]
\centering
\caption{Intra-Cluster Consensus Execution}
\label{alg:intra_cluster}
\small
\begin{algorithmic}[1]
\Require Subtask $s$, cluster $C$, refinement rounds $k$, weights vector $w$
\Ensure Final consensus answer $\text{ans}_C$
\State Initialize local answers $\text{ans}_i \gets \text{draft}_{a_i}$ for each $a_i \in C$
\For{$r = 1$ to $k$}
  \For{each $a_i \in C$ in parallel}
    \State Broadcast $\text{ans}_i$ to all peers and receive their answers
    \State $\text{ans}_i \gets \texttt{UPDATE}(\{\text{ans}_j : a_j \in C\}, w)$ \Comment{weighted aggregation/voting}
  \EndFor
  \If{$\texttt{TASK\_COMPLETE} ~ \text{in all} ~ \{\text{ans}_j : a_j \in C\}$} 
    \State \textbf{break}
  \EndIf
\EndFor
\State Choose representative answer $\text{ans}_C$ (e.g., by majority or highest weight)
\State \Return $\text{ans}_C$
\end{algorithmic}
\end{algorithm}

The complexity of the intra-cluster protocol depends on the cluster size $|C|$ and the number of rounds $k$. Each round involves an all-to-all exchange within the cluster followed by a local aggregation, giving $O(k\,|C|^2)$ communication steps. In practice, we limit cluster sizes to 3-5 agents to balance diversity and overhead, avoiding significant costs.


\paragraph{Phase~5: Reporting and DAG updates.} After consensus, each cluster emits a \texttt{TASK\_COMPLETE} message containing the refined answer. The orchestrator subscribes to these messages on the appropriate MQTT channel and updates the DAG state, marking subtask $s$ as complete. If consensus cannot be reached within the allotted rounds $k$, the orchestrator either reassigns the subtask or accepts the highest-scoring draft, mirroring fallback behaviours in classical multi-agent planning. Reporting is a constant-time operation per subtask and therefore does not dominate the overall complexity.

\paragraph{Phase~6: Result synthesis and merging.} Once all subtasks have reported completion, the orchestrator synthesises the final answer by traversing the DAG $G$ in topological order and merging the outputs of predecessors with the current subtask result. Algorithm~\ref{alg:synthesis} describes this synthesis.

\begin{algorithm}[!ht]
\centering
\caption{Result Synthesis and Merging}
\label{alg:synthesis}
\small
\begin{algorithmic}[1]
\Require DAG $G=(\mathcal{S},E)$, answers $\{\text{ans}_{s_i} : s_i \in \mathcal{S}\}$
\Ensure Final solution $\text{ans}_t$
\State $\text{ans}_t \gets \emptyset$
\For{each $s_i$ in topological order}
  \State $\text{pre} \gets \{\text{ans}_{s_j} : (s_j \to s_i) \in E\}$ \Comment{predecessor answers}
  \State $\text{combined} \gets \text{SYNTH}(\text{pre} \cup \{\text{ans}_{s_i}\})$ \Comment{merge operator}
  \State $\text{ans}_{s_i} \gets \text{combined}$
\EndFor
\State Set $\text{ans}_t \gets \text{ans}_{s_{root}}$ or merge sink nodes
\State \Return $\text{ans}_t$
\end{algorithmic}
\end{algorithm}

Topological traversal and merging runs in $O(|\mathcal{S}| + |E|)$ time, as each subtask is processed exactly once and the merge operator combines a bounded number of predecessor answers. This final phase ensures that dependencies are respected and partial results are propagated forward without blocking unrelated branches, realizing the fine-grained concurrency promised by the DAG model.

\paragraph{Summary of complexity.}
Bringing the phases together, the overall complexity of the FoA execution pipeline for a single task is
\begin{align}
&O\Bigl(n d + m |\mathcal{S}| \log |\mathcal{S}|\Bigr)+ \qquad \qquad &\rhd \text{Dynamic Task Decomposition}\\
&O\Bigl(\sum_{s \in \mathcal{S}} \sum_{a_i \in \mathcal{A}_s}(r_i + t_i)\Bigr)+  \qquad \qquad &\rhd \text{First Draft Generation} \\
&O\Bigl(\sum_{s \in \mathcal{S}} |\mathcal{A}_s|^2 (d + \log |\mathcal{A}_s|)\Bigr)+ \qquad \qquad &\rhd \text{Semantic Cluster Formation} \\
&O\Bigl(\sum_{C \in \mathcal{C}} k|C|^2\Bigr)+ \qquad \qquad &\rhd \text{Cluster Consensus} \\
&O(|\mathcal{S}| + |E|) \qquad \qquad &\rhd \text{Synthesis}
\end{align}

where the terms correspond respectively to dynamic decomposition and proposal merging, initial draft generation, cluster formation, intra-cluster consensus, and final synthesis. In practice, many of these phases execute in parallel, and cluster sizes remain small, yielding efficient end-to-end execution. Our design thus achieves the adaptability and flexibility promised by dynamic task decomposition and multi-agent collaboration.





\section{MQTT Implementation Details}

\paragraph{MQTT Topic Hierarchy.} FoA uses a structured topic namespace:
\begin{itemize}
\item \texttt{foa/orchestrator/jobs} - Job submissions
\item \texttt{foa/agents/\{agent\_id\}/tasks} - Individual agent tasks
\item \texttt{foa/clusters/\{cluster\_id\}/channel} - Intra-cluster communication
\item \texttt{foa/capabilities/updates} - VCV updates
\item \texttt{foa/policies/enforcement} - Policy compliance events
\end{itemize}

\paragraph{Embedding Pipeline.} VCV embeddings use a two-stage process:
\begin{enumerate}
\item Raw capability descriptions processed through a specialized tokenizer
\item 768-dimensional embeddings generated via Nomic Embed~\cite{nussbaum2024nomic}
\item L2 normalization applied for cosine similarity computation
\item Optional dimensionality reduction to 256 for storage efficiency
\end{enumerate}

\section{Additional Experimental Results}\label{sec:add-exp}

Our evaluation compares FoA against several baselines. For each task, we instantiate four \emph{Agent-1} models with Specs covering a broad spectrum of general medicine; each is a large language model fine-tuned from a foundation model using Group Relative Policy Optimisation (GRPO) on its domain data~\cite{shao2024deepseekmath}. Baselines include (i) the best individual agent (selected according to its standalone HealthBench Hard score), (ii) an \emph{uncoordinated ensemble} that averages the four agents' responses without decomposition or clustering, and (iii) \emph{random assignment}, where subtasks are assigned to agents uniformly at random without consulting their capability vectors. All systems enforce the same AI safety policies via a Bloom filter that blocks prompts containing unsafe content.

We configure FoA as follows. The orchestrator maintains a sharded HNSW index over 256-dimensional Versioned Capability Vectors and performs semantic routing with a similarity threshold $\alpha_a=0.3$ (\texttt{FOE\_DECOMP\_THRESHOLD}). At most four agents are considered to propose decompositions (\texttt{FOE\_DECOMP\_MAX\_AGENTS}), and each task is split into between two and four subtasks (\texttt{FOE\_DECOMP\_SUBTASKS\_MIN/MAX}). Candidate decompositions are merged when their subtask embeddings have a cosine similarity of at least $0.5$ (\texttt{FOE\_DECOMP\_MERGE\_SIM}). Clustering is enabled and uses EMQX~v5~\cite{emqx2024harnessing} as the MQTT broker. Agents in a cluster communicate over a dedicated topic for a maximum of $k=3$ refinement rounds; clusters are restricted to one to four agents with a formation threshold of $0.2$ (\texttt{FOE\_CLUSTER\_SIM\_THRESHOLD}), allowing A-1 agents to participate in more than one cluster with a single job timeout of $300$~s (Fig.~\ref{fig:clust}). Agent processes run on \textbf{one} NVIDIA GPU optimizing for latency and energy consumption using a 4-bit quantized small language model from a pool of four classes of pre-aligned small language models: \textsc{gemma3}, \textsc{qwen3}, \textsc{deepseek-r1}, \textsc{gpt-oss}. In our experiments, we considered only small language models with a maximum of 20B parameters that could run on consumer-scale hardware using the Ollama API. In this particular experiment, we set medgemma3~\cite{sellergren2025medgemma} as the default pick from the gemma3 family. We used \textsc{medgemma:27b} as the grader used for HealthBench evaluation on a separate cluster, and we report the mean and bootstrap standard deviation over three random seeds. Responses are generated for all 1,000 HealthBench Hard conversations and evaluated using the official \texttt{simple-evals} pipeline.

\paragraph{Assets Used}
The language models considered in this work include \textsc{gemma3}, \textsc{qwen3}, \textsc{deepseek-r1}, and \textsc{gpt-oss}. \textsc{gemma3} is distributed under Google's Gemma Terms of Use, \textsc{qwen3} under the Apache License~2.0, \textsc{deepseek-r1} under the MIT License, and \textsc{gpt-oss} under the Apache License~2.0. The OpenAI \textsc{HealthBench} benchmark (dataset) and the accompanying reference evaluation code are made available under the MIT License.

\paragraph{Used Computer Resources}
All experiments were conducted on a workstation equipped with an \textbf{Intel\textsuperscript{\textregistered} Xeon\textsuperscript{\textregistered} w5-2455X CPU} (64-bit, AVX-512 enabled) and \textbf{63 GB of system memory}. Computations were accelerated using an \textbf{NVIDIA\textsuperscript{\textregistered} RTX\texttrademark{} A4000 GPU} with \textbf{16 GB of GDDR6 memory} (CUDA~12.8, driver~573.53) for an average runtime of $536.7$ seconds per query. The system operated in a virtualized environment with Red Hat VirtIO drivers for storage and file system access, and networking provided through a \textbf{10~Gbit/s Ethernet interface}.

\section{MQTT-Based Workflow}

Motivated by CAFEIN\textsuperscript{\textregistered}~\cite{cafein}, FoA employs MQTTv5 as its communication backbone for orchestration. This protocol offers lightweight publish/subscribe semantics together with Quality of Service guarantees, making it well-suited for coordinating large sets of heterogeneous agents. Instead of relying on direct RPC or HTTP calls, interactions are expressed through a structured hierarchy of topics that naturally reflects the lifecycle of a task. The MQTT broker manages message passing and communication between nodes in the federated network, while also handling \emph{authentication} and \emph{authorization} to ensure secure and trusted interactions. MQTTv5 was chosen over alternative application protocols, such as HTTP, for its ability to efficiently support one-to-many asynchronous communication, while providing embedded security and strong scalability features.

At the beginning of the process, new jobs are published to the orchestrator under a dedicated submission namespace. Agents simultaneously advertise their current capabilities by emitting versioned capability vectors on a separate update channel. These updates are disseminated using a $\Delta$-gossip protocol and remain consistent across the federation due to MQTT's delivery guarantees. The orchestrator consumes both streams - tasks and capabilities - and decomposes each job into subtasks, which are then routed to the appropriate agent-specific topics. 

Whenever a task requires collaboration among multiple agents, the orchestrator instantiates a cluster channel. Within this temporary communication space, agents exchange intermediate results, critiques, and refinements until a consensus signal is produced. Policy enforcement events are broadcast in parallel on a separate namespace, ensuring that compliance checks and auditability are embedded in the same communication fabric as task execution. Finally, results are published to the global result channel, where the orchestrator merges them according to the dependency structure of the job's directed acyclic graph. 

This design ensures that every stage of the workflow - submission, capability dissemination, assignment, collaborative execution, enforcement, and result collection - is mediated by MQTT topics. The resulting system benefits from reliability and auditability, while maintaining the scalability and low overhead that are required in heterogeneous and bandwidth-constrained environments.
\end{document}